\documentclass[preprint,5p,times]{elsarticle}

\usepackage{amssymb}
\usepackage{graphicx}

\usepackage{hyperref}
\hypersetup{
  colorlinks=true,
}

\journal{arxiv}

\begin{document}
\hypersetup{
  linkcolor=red,
  urlcolor=blue,
  citecolor=red
}

\begin{frontmatter}



\title{GPT as Knowledge Worker: \\A Zero-Shot Evaluation of (AI)CPA Capabilities}

\twocolumn
\author[a]{Jillian Bommarito}
\ead{jill@273ventures.com}
\author[a,b,c,d]{Michael J Bommarito II}
\author[a]{Jessica Katz}
\author[a,b,c,d]{Daniel Martin Katz}
\affiliation[a]{273 Ventures, LLC}
\affiliation[b]{Illinois Tech - Chicago Kent College of Law}
\affiliation[c]{Bucerius Law School}
\affiliation[d]{CodeX - The Stanford Center for Legal Informatics}

\begin{abstract}
The global economy is increasingly dependent on knowledge workers to meet the needs of public and private organizations. While there is no single definition of knowledge work, organizations and industry groups still attempt to measure individuals' capability to engage in it. The most comprehensive assessment of capability readiness for professional knowledge workers is the Uniform CPA Examination developed by the American Institute of Certified Public Accountants (AICPA). In this paper, we experimentally evaluate OpenAI's \textsc{text-davinci-003} and prior versions of GPT on both a sample Regulation (REG) exam and an assessment of over 200 multiple-choice questions based on the AICPA Blueprints for legal, financial, accounting, technology, and ethical tasks.  First, we find that \textsc{text-davinci-003} achieves a correct rate of 14.4\% on a sample REG exam section, significantly underperforming human capabilities on quantitative reasoning in zero-shot prompts.  Second, \textsc{text-davinci-003} appears to be approaching human-level performance on the Remembering \& Understanding and Application skill levels in the Exam absent calculation.  For best prompt and parameters, the model answers 57.6\% of questions correctly, significantly better than the 25\% guessing rate, and its top two answers are correct 82.1\% of the time, indicating strong non-entailment. Finally, we find that recent generations of GPT-3 demonstrate material improvements on this assessment, rising from 30\% for \textsc{text-davinci-001} to 57\% for \textsc{text-davinci-003}.  These findings strongly suggest that large language models have the potential to transform the quality and efficiency of future knowledge work.
\end{abstract} 

\end{frontmatter}

\section*{Introduction}
\label{sec:introduction}
Knowledge work is an increasingly important segment of the global economy, with qualified professionals providing services in areas such as law, finance, accounting, economics, and technology.  Leading management theorists began exploring definitions of ``knowledge workers'' and approaches for their training nearly seven decades ago \cite{drucker1959landmarks, bloom1956taxonomy, drucker1967effective}.  Since then, the percentage of the population that ``thinks for a living'' has grown dramatically.  As of 2021, the Big 4 - Deloitte, EY, PWC, and KPMG - alone employ over one million people \cite{statista2021big4}; some definitions of knowledge work suggest that the true number of knowledge workers is in the hundreds of millions or even billions \cite{cortada2009rise}.

As their roles and activities may generate substantial value - and liability - many organizations require these knowledge workers to demonstrate their preparedness through comprehensive assessments, such as the so-called CPA, CFA, or Bar exams.  While there is no universally-accepted definition of knowledge work \cite{alvesson2001knowledge}, public accounting is a multidisciplinary practice that requires legal, financial, accounting, auditing, technology, and ethical knowledge and skills - all domains clearly within the scope of knowledge work.  As the test used to assess the readiness of candidates for this profession, the American Institute of Certified Public Accountants (AICPA) Uniform CPA Examination (``CPA Exam'' or ``Exam'') is the most comprehensive, well-known assessment of knowledge work readiness \cite{yeaton2020cpa}.  As compared to other assessments or examinations, the CPA Exam is broader, more practice-based, and more regularly updated to meet the changing landscape.  This trend is perhaps best demonstrated by the fact that the commercial organizations most associated with the AICPA - the Big 4 - have accumulated practically every type of knowledge work under their umbrella, including even cybersecurity and traditional legal services  \cite{herda2021big4, donelson2020revival, banham2017cybersecurity}.

The AICPA and the National Association of State Boards of Accountancy have undertaken a joint effort to ensure that the CPA licensure model reflects the ``rapidly changing skills and competencies the practice of accounting requires today and will require in the future'' \cite{aicpa2022maintaining}. The Exam is produced by the AICPA based on input from stakeholders in the professional services industry, academia, and governmental agencies.  The Exam has been continually updated to meet changing regulations, standards, technology, and market expectations for over 100 years \cite{yeaton2020cpa, king2017accounting}.  While the Exam continues to evolve \cite{king2017accounting, kravitz2012evaluating}, it was historically adapted from the best-known educational framework, Bloom's cognitive taxonomy \cite{bloom1956taxonomy}, to organize the assessment of practical, professional requirements into four skill levels \cite{aicpa2021blueprints}.  Though the exam will undergo significant structural changes in 2024, the current implementation of the exam has been divided into four sections: Auditing and Attestation (AUD), Business Environment and Concepts (BEC), Financial Accounting and Reporting (FAR), and Regulation (REG).  These four sections cover concepts, laws, rules, and relationships in legal, financial, accounting, and technology domains, common denominators among many knowledge professions.\footnote{Interested readers should review Table \ref{tab:exam_blueprint_areas} for the list of all concept areas.}

Previous decades of research into artificial intelligence (AI) have not yielded general models capable of performing knowledge work.  While point solutions in many legal, financial, or accounting domains have shown value or reached adoption, there has been no demonstration of AI that can span multiple task types in professional services.  This gap can likely be attributed to multiple reasons, including the breadth and depth of knowledge required to be indexed and recalled, as well as the complexity of translating this knowledge into work product in the context of realistic client engagements.  To make matters more difficult, professional services like accounting, finance, and law also often require a combination of quantitative and qualitative skills. 

Recent research has, however, shown potential to address at least some of these capability gaps.  Advances in natural language processing (NLP), machine learning (ML), and computing over the last decade have produced material improvements in state-of-the-art performance on linguistic tasks that require deeper semantic understanding or feature more complex syntax \cite{Mikolov2013} \cite{Pennington2014} \cite{Peters2018}.  More importantly, some types of models have begun to demonstrate the ability to address dramatically different task types, sometimes even in zero-shot use cases where there is no additional fine-tuning or customization.  While neural network research is not new \cite{neumann1958computer} \cite{Rumelhart1986}, the rate of progress has increased dramatically since 2013, and, in particular, transformer-based architectures \cite{Vaswani2017} have been shown to produce previously-unseen capabilities to generalize across tasks \cite{Devlin2018} \cite{Brown2020} \cite{Zaheer2020} \cite{Scao2022} \cite{Thoppilan2022}.  

The most accessible and well-known of these transformer-based models is OpenAI's family of large language models known as Generative Pre-trained Transformer or ``GPT'' \cite{Brown2020} \cite{radford2018improving}.  The latest versions of GPT, often referred to as GPT-3 or GPT-3.5, are proprietary large language models, and these models are only available to OpenAI customers.  One benefit of this approach is that it provides an important layer of legal and ethical moderation, as well as simplifying the user experience, such as by preprocessing input text or images.  As of this publication, the OpenAI provides API endpoints for text completion, code completion, image generation, and embedding generation tasks.  OpenAI has also recently unveiled ChatGPT, a public-facing ``chatbot'' built on GPT-3.5, which reportedly generated over 1M user sign-ups within just a few days of release. 

As GPT-3 and its derivatives are proprietary machine learning models in production within a reinforcement learning platform, we cannot precisely describe them.  However, based on GPT-3’s original publication in July 2020 and subsequent material, these models are likely derived from an autoregressive language model with 175 billion parameters, 96 layers, and a batch size of 3.2M. OpenAI has launched or published a number of GPT-3 derivative models, most notably InstructGPT-3 and Codex 12B, which are colloquially referred to as GPT-3.5. The most advanced model in production in its API is \textsc{text-davinci-003}, an improvement on \textsc{text-davinci-002}, which is an InstructGPT model based on \textsc{code-davinci-002}, a base model for pure code-completion tasks, per OpenAI documentation.  Our results in this paper are primarily based on \textsc{text-davinci-003}, as detailed in Section \ref{sec:methods}, though we also include results from older models for comparison and forecasting.

While \textsc{text-davinci-003} and ChatGPT have demonstrated state-of-the-art performance on a wide range of tasks in zero-shot and few-shot contexts, there was previously little reason to believe that these models could perform even reasonably well in general assessments across the domains of finance, law, and accounting.  However, in recent prior work on the Bar Exam \cite{bommarito2022gpt}, the authors have shown that \textsc{text-davinci-003} could achieve near-parity with human test-takers in two of seven sections of the Multistate Bar Exam (MBE); more strikingly, generation-over-generation model performance suggests that an LLM like GPT-3.5 may be capable of passing the Bar Exam in the near future.

While the Bar Exam offered one measure of performance for GPT-3.5, it is arguably not the ideal instrument to evaluate readiness for multidisciplinary knowledge work.  As noted, the CPA Exam requires a wider range of knowledge, including not only law, but also finance, accounting, technology, and ethics.  Therefore, in order to evaluate whether and how current state-of-the-art models in AI might be applied to knowledge work, we experimentally evaluate the performance of ``GPT as knowledge worker'' through the skills and concepts outlined in the CPA Exam.  Our analysis suggests both areas where GPT-3.5 may be useful today and areas where substantial research and development is still required.

\section*{AICPA Exam}
The Uniform CPA Examination is a modern, computerized assessment based on psychometric and statistical techniques.  While prior paper-based generations of the Exam might have been compared to traditional linear exams, the current Exam is a dynamic, adaptive exam \cite{melican2009designing}, best compared to exams like the current GRE or GMAT. Linear exams present the test-taker with a preset sequence of test questions, while dynamic exams adapt to each test-taker in response to the answers provided in prior questions. 

\begin{table}[htpb]
    \small
    \centering
    \begin{tabular}{|c|c|}
    \hline
    \textbf{Section} & \textbf{Student Pass Rate}\\\hline
        AUD & 48.7\%\\\hline
        BEC & 59.7\%\\\hline
        FAR & 44.9\%\\\hline
        REG & 61.1\%\\\hline
    \end{tabular}
    \caption{Passage rates of students in 2022 as reported by the AICPA \cite{aicpa2023homepage}.}
    \label{tab:student_performance_by_section}
\end{table}

\pagebreak
\onecolumn
\begin{table}[h]
\small
\centering
\begin{tabular}{ |p{2.6cm}|p{10cm}| }
 \hline
 \textbf{Skill Level} & \textbf{Description}\\
 \hline
 Evaluation & The examination or assessment of problems, and use of judgment to draw conclusions.\\
 \hline
 Analysis & The examination and study of the interrelationships of separate areas in order to identify causes and find evidence to support inferences.\\
 \hline
 Application & The use or demonstration of knowledge, concepts, or techniques.\\
 \hline
 {Remembering \&\newline Understanding} & The perception and comprehension of the significance of an area utilizing knowledge gained.\\
 \hline
\end{tabular}
\caption{AICPA Uniform CPA Examination Skill Levels}
\label{tab:blueprint_skills}
\end{table}

\begin{table}[h]
    \small
    \centering
    \begin{tabular}{|p{2.6cm}|p{1.3cm}|p{2.25cm}|p{6cm}|}
    \hline
    \textbf{Skill} & \textbf{Area} & \textbf{Content} & \textbf{Task} \\
    \hline
    Remembering \& Understanding & Internal Controls & Sarbanes-Oxley Act of 2002 &  Identify and define key corporate governance provisions of the Sarbanes-Oxley Act of 2002. \\ \hline
    Application & Internal Controls & Sarbanes-Oxley Act of 2002 & Identify regulatory deficiencies within an entity by using the requirements associated with the Sarbanes-Oxley Act of 2002. \\ \hline
    \end{tabular}
    \caption{Example AICPA Uniform CPA Examination Tasks}
    \label{tab:example_skills}
\end{table}

\begin{table}[h]
    \small
    \centering
    \begin{tabular}{|p{10.5cm}|}
        \hline
        \textbf{Auditing and Attestation (AUD)} \\
        \hspace{1em} Ethics, Professional Responsibilities and General Principles \\
        \hspace{1em} Assessing Risk and Developing a Planned Response \\ 
        \hspace{1em} Performing Further Procedures and Obtaining Evidence \\ 
        \hspace{1em} Forming Conclusions and Reporting\\
        \textbf{Business Environment and Concepts (BEC)} \\
        \hspace{1em} Enterprise Risk Management, Internal Controls and Business Processes \\ 
        \hspace{1em} Economics \\ 
        \hspace{1em} Financial Management \\ 
        \hspace{1em} Information Technology \\ 
        \hspace{1em} Operations Management \\
        \textbf{Financial Accounting and Reporting (FAR)} \\
        \hspace{1em} Conceptual Framework, Standard-Setting and Financial Reporting \\
        \hspace{1em} Select Financial Statement Accounts \\
        \hspace{1em} Select Transactions \\
        \hspace{1em} State and Local Governments \\
        \textbf{Regulation (REG)} \\
        \hspace{1em} Ethics, Professional Responsibilities and Federal Tax Procedures \\
        \hspace{1em} Business Law \\
        \hspace{1em} Federal Taxation of Property Transactions \\
        \hspace{1em} Federal Taxation of Individuals \\
        \hspace{1em} Federal Taxation of Entities \\
        \hline
    \end{tabular}
    
    \caption{Uniform CPA Examination Blueprints - Content Areas}
    \label{tab:exam_blueprint_areas}
\end{table}

\twocolumn

The Examination is divided into four sections that test-takers sit for independently: Auditing and Attestation (AUD), Business Environment and Concepts (BEC), Financial Accounting and Reporting (FAR), and Regulation (REG).  Each section of the Exam is divided up into at least four testlets that feature scenarios, multiple choice questions, calculated amounts, short answer, and related evidence and research material.  The passage rates of Exam sections are presented in Table \ref{tab:student_performance_by_section}; the AICPA does not publish statistics related to per-question or per-section test-taker accuracy.

By its very design, the Exam is meant to be a practical assessment of real-world tasks and requisite skills \cite{aicpa2022maintaining, melican2009designing}.  It rigorously assesses candidates on their readiness across a broad range of concepts and skill levels progressing through (i) Remembering \& Understanding, (ii) Application, (iii) Analysis, and (iv) Evaluation.

The overall design of the Exam is best viewed through the Uniform CPA Examination Blueprints (``Blueprints'') \cite{aicpa2021blueprints}, which document how concepts and tasks are adapted from Bloom’s taxonomy of the cognitive domain \cite{bloom1956taxonomy}.  An overview of the Exam and sample skills and tasks are provided in Tables \ref{tab:blueprint_skills}, \ref{tab:example_skills}, and \ref{tab:exam_blueprint_areas}.  The Blueprints are regularly updated by the AICPA and are the most detailed, representative outline of the test's construction.

Importantly, many of the tasks detailed in the Blueprints include an element of arithmetic.  For example, many questions that include workpapers or sample financial statements expect the test-taker to first determine which numbers to include or exclude in arithmetic expressions, then to evaluate the resulting expression to calculate a specific amount.  Sometimes, these expressions are as simple as $A = L + E$, but in many cases, they involve more complex expressions based on tables with dozens of numbers and related materials.  Based on prior research and experience with LLMs, we strongly suspected that GPT-3.5 would struggle with zero-shot quantitative reasoning in this context.

\section*{Data}
\label{sec:data}
While there is an active body of research on quantitative reasoning with fine-tuning or few-shot contexts \cite{dai2022, qian2022limitations, sharma2022overcoming, muffo2022evaluating}, we constrain our results in this study to zero-shot prompts to better assess the ``intrinsic'' capability of these models.  Therefore, we prepared two separate assessments to allow us to isolate the arithmetic or quantitative capabilities from other elements of the Exam.

\subsection*{Assessment 1: Sample Exam - Regulation}
\label{sec:realistic_exam}
The first assessment is intended to approximate the real Uniform CPA Examination using the AICPA’s online, publicly-available sample exams.  These tests “include two multiple-choice testlets and three task-based simulation testlets for [...] Auditing and Attestation (AUD), Financial Accounting and Reporting (FAR) and Regulation (REG);” the fourth section, BEC, is shorter.  Between AUD, FAR, and REG, we utilize the REG section as it contains the most balanced distribution of skill types and quantitative and qualitative reasoning.  Therefore, a test session of the REG exam as provided on the AICPA’s site was transcribed on January 3rd, 2023, including correct answers.  All questions are formatted as simple text or, where evidence or workpapers are formatted in tables or lists, as Markdown.

This process results in 40 test questions across five testlets.  Two of these five testlets consist of multiple-choice questions, with a total of 15 questions ranging from four to six options each.  Of the remaining 25 questions, 24 require the test-taker to indicate the correct financial amount and one requires the test-taker to research authoritative material made available within the exam.  While we cannot redistribute these test questions directly, interested readers can directly access and take the AICPA’s online sample exams at no cost.

A partially-redacted sample question from this assessment is provided for reference below:
\vspace{.3cm}

\begin{center}
\small\textbf{Assessment 1: Sample Question}
\small
\begin{verbatim}
  Question: All taxpayers file their Form 1040 using
  the tax filing status of single. Assume that [...].

  Situation:
  $6,000 - Loss on sale of [...]
  $10,000 - Contribution to the capital [...]
  $3,000 - Write-off of a worthless [...]

  What is the taxpayer’s adjusted gross income?
  Answer: $65,000  
\end{verbatim}
\normalsize
\end{center}

\vspace{.3cm}

\subsection*{Assessment 2: Synthetic MCQ Assessment}
\label{sec:knowledge_application_assessment}
As noted above, the Uniform CPA Examination is organized around Bloom’s cognitive taxonomy \cite{bloom1956taxonomy}, which is a widely-adopted framework for structuring learning objectives and capabilities.  The taxonomy is generally conceptualized as a pyramid divided into six levels: Knowledge, Comprehension, Application, Analysis, Synthesis, and Evaluation or Creation. As noted above in Table \ref{tab:blueprint_skills}, the AICPA has adapted these skill levels into four simpler groups.  The top two levels - Evaluation and Analysis - not only most frequently feature arithmetic, but in practice, are also frequently the most nuanced, contextual tasks that real professionals address.  

As an example, tasks like ``Evaluate the reasonableness of significant accounting estimates [...]'' are ones for which, for legal and ethical reasons, human oversight will likely remain necessary.  

Therefore, we focused this second assessment on the foundational levels of the AICPA's skill pyramid - Remembering \& Understanding and Application. To do so, we reviewed every task in the AICPA's Blueprints, dated October 18, 2021, to identify all relevant tasks.  For each task, the lead author, a CPA, prepared at least one question to address each task and skill level identified.  In sections where there were fewer than 50 relevant Blueprint tasks, we randomly sampled tasks and added additional questions to ensure that all sections had at least 50 samples.  While this means that the calculation of overall accuracy rate overweights sections such as BEC, we are not focused on test passage \textit{per se} in this research and therefore prefer breadth and power.

These questions have been prepared, to the best of our abilities, to mimic the nature and difficulty of real questions on the Exam.  In addition to reviewing material provided by the AICPA itself, the authors also reviewed material and sample questions prepared by McGraw-Hill Education and Becker Professional Education to ensure that our test questions were at least as difficult and broad as theirs.  All questions were drafted solely by the authors, and a  sample question from each section of this assessment is provided for reference below.\\

\begin{center}
\small
\textbf{Assessment 2: Synthetic REG Question}
\begin{verbatim}
  Question: Which of the following types of contract 
  does not require a written element in order to be 
  enforceable?

  A. Contracts for the sale of goods for $500 
     or more
  B. Contracts to act as surety
  C. Contracts for the sale of a house
  D. Contracts for leases of land for less than
     one year

  Answer: D
\end{verbatim}
\end{center}

\vspace{.5cm}

\begin{center}
\small
\textbf{Assessment 2: Synthetic BEC Question}
\begin{verbatim}
  Question: Which of the following elements is not 
  part of the formula for calculating the cost of 
  retained earnings using the Capital Asset Pricing
  Model?

  A. The risk-free rate
  B. The pre-tax cost of long-term debt
  C. The company’s beta coefficient
  D. The market risk premium

  Answer: B
\end{verbatim}
\end{center}

\vspace{.5cm}

\begin{center}
\small
\textbf{Assessment 2: Synthetic FAR Question}
\begin{verbatim}
  Question: Which of the following investment 
  types is eligible to be reported in the 
  financial statements at amortized cost?

  A. Available-for-sale equity securities
  B. Available-for-sale debt securities
  C. Held-to-maturity debt securities
  D. Trading equity securities

  Answer: C
\end{verbatim}
\end{center}

\vspace{1cm}

\begin{center}
\small
\textbf{Assessment 2: Synthetic AUD Question}
\begin{verbatim}
  Question: Which of the following disclosures 
  related to the fair value of investments in 
  securities is required for a nonissuer?

  A. Purchases and issuances for each class of
     investments
  B. Rollfoward of recurring level 3 fair value
     measurements
  C. Disclosures for financial instruments not 
     measured at fair value
  D. The range and weighted average of 
     significant unobservable inputs

  Answer: A
\end{verbatim}
\end{center}

\vspace{.3cm}

\normalsize
These questions, like natural language in the law itself, can be subject to pedantic interpretation; for example, in the Auditing and Attestation (\textbf{AUD}) question above, an experienced practitioner might qualify choice B by stating that it depends on whether it's a ``full rollforward'' or a limited number of separate elements of the rollforward. Similar to the actual CPA Exam, some of our questions may require the selection of the ``best'' option.

\normalsize

In total, we produced 208 questions across the four sections of the Exam.  The distribution of these questions is detailed in Table \ref{tab:questions_per_section} below. All questions are available in the online SI on GitHub.  Like the AICPA's exam designers themselves, we expect that there will be issues with the design or scoring of our questions, and we encourage readers to submit additional questions or suggested clarifications via corresponding email or GitHub.  As errata may be detected or new questions accepted, updated results may be available in the online SI.

\begin{table}[htpb]
    \small
    \centering
    \begin{tabular}{|c|c|c|}
    \hline
    \textbf{Assessment} & \textbf{Section} & \textbf{Number of Questions}\\ \hline
        1 & REG & 40 \\ \hline
        2 & AUD & 54  \\ \hline
        2 & BEC & 50 \\ \hline
        2 & FAR & 51 \\ \hline
        2 & REG & 53 \\ \hline

    \end{tabular}
    \caption{Number of AICPA and author-prepared questions per section.}
    \label{tab:questions_per_section}
\end{table}

\section*{Methods}
\label{sec:methods}
In prior work on the Bar Exam \cite{bommarito2022gpt}, we outlined a method for experimentally evaluating OpenAI's models. For multiple choice question (MCQ) assessments in this paper, we follow this approach  as closely as possible; calculated amounts and short answers are compared to the correct answer after stripping and reformatting answers.  For example, $(10,000)$, $(10000)$, and $-10,000$ are identical in the automated scoring of the model's responses.\footnote{Parentheses are used as shorthand in the accounting industry for negative amounts.}

As in prior research, our evaluation is based on generating zero-shot prompts for the \textsc{text-davinci-003} text completion API.  Unlike in our prior research \cite{bommarito2022gpt}, we are able to fully open-source the source code and questions created in Assessment 2.  While replication of results requires an OpenAI account and accepting the AICPA's terms of use, we have again attempted to provide researchers with as much replication detail as is possible under the circumstances.  

\subsection*{Prompt Engineering and Responses}
Our ability to understand these large language models is constrained both by our limited scientific understanding and the proprietary nature of OpenAI's models \cite{bommarito2022gpt}.  Despite this gap, many have documented that such models are unexpectedly sensitive to the specific prompts they are provided.  The practice of writing such prompts is typically referred to as ``prompt engineering,'' and details of prompt engineering are critical to replication of studies involving LLMs.  

In this research, we experimented with answer types, contextualization, and justification in prompt engineering \cite{kirk2022improving}.  The following prompt variations were tested in at least one sample, although variations between Assessment 1  and Assessment 2 are required due to question types. For Assessment 1, the prompts define entailment or recall tasks, i.e., where the model must select the correct or most correct answer, as well as open-ended problems where the model must calculate the correct monetary amount.  For Assessment 2, all questions are designed to evaluate traditional entailment tasks.  Complete details are available in the source and data in the online SI.

\begin{enumerate}
    \item Answer.  Ask the model to answer with:
    \begin{itemize}
      \item its best choice only.
      \item its best and worst choices.
      \item its top three rank-ordered choices.
    \end{itemize}
    \item Contextualization.  Ask the model to imagine it is:
    \begin{itemize}
        \item taking the CPA exam.
        \item designing the CPA exam.
        \item an accountant in the United States.
        \item a tax professional in the United States.
        \item a legal professional in the United States.
        \item a Big 4 accountant in the United States.
        \end{itemize}
     \item Justification. Require the model to provide:
     \begin{itemize}
        \item an explanation of its choices.
        \item an explanation and citation to authority or source.
        \item an explanation and citation within a specific list of authorities or sources.
    \end{itemize}
\end{enumerate}

Generated prompts are combined with questions and sent to the OpenAI API endpoint.  The prompt and complete JSON response, including the OpenAI API request ID, are logged for all questions for all assessments.  The API response is parsed and stored for scoring, qualitative analysis, and open source release.  For scoring, no responses were manually altered or evaluated by humans.

In general, most prompts produced similar performance, clustering near the central tendency of 55\% noted in Table \ref{tab:performance_by_model_generation}.  In a number of cases, contextualization or justification resulted in models that performed better on one section but worse on another section. Contextual variations suggest differences in the nature of advice between professions. Justification variations suggest differences in the complexity or state of codification across subject areas. Additional details, complete responses, and details regarding phenomena such as hallucination are provided in the SI.

\subsection*{Model (hyper)parameters}
As the AICPA curriculum itself notes, many models are sensitive to small changes in their inputs, and LLMs are no different.  In addition to prompt sensitivity, they are often highly sensitive to the parameters set in training and inference.  While our ability to intepret results or identify all (hyper)parameters is limited by the proprietary nature of GPT, we did evaluate how altering some model parameters impacts the performance of the model.  We do not vary the maximum token output or attempt nucleus sampling; however, we do evaluate the following parameters for at least one prompt:
\begin{enumerate}
  \item \textsc{temperature}: Sampling temperature; 0.0 is deterministic, higher is more ``random.''  We tested values in \{0.0, 0.5, 1.0\}.
  \item \textsc{best\_of}: ``Generates [N] completions server-side and returns the ``best'' (the one with the highest log probability per token).''  We tested values in $\{1, 2, 4\}$.
\end{enumerate}

\subsection*{Fine-tuning and Historical Models}
While OpenAI does provide an API for fine-tuning models including \textsc{text-davinci-003}, this publication is focused on the zero-shot performance of the model itself.  Furthermore, based on prior experience in similar problems \cite{bommarito2022gpt}, we do not believe that fine-tuning text completion at small sample sizes would improve the models' performance.  In some circumstances, others have found success in subsequent supervised or unsupervised re-training of some or all layers of an LLM \cite{Dunn2022}\cite{Huang2022}, while others have documented circumstances in which fine-tuning results in unexplained model degradation.  In our prior work \cite{bommarito2022gpt}, we noted a significant decrease in fine-tuned \textsc{text-davinci-003} performance at the scale of our training data.  While it is possible that this performance decrease is explained by the 50\% head layer contraction required by OpenAI's API, we are unable to test further without access to details of fine-tuning or resulting weights.

In addition to \textsc{text-davinci-003}, OpenAI also makes a number of other models available through its API, including smaller and older iterations of the GPT family.  We repeated our testing with the \textsc{text-davinci-001}, \textsc{text-curie-001}, \textsc{text-babbage-001}, and \textsc{text-ada-001} models provided through the OpenAI API.

\section*{Results}
In total, across all prompts and parameters tested, we asked \textsc{text-davinci-003} to answer over 50,000 questions in more than 700 independent assessment sessions.  Details of the number of sessions and parameter values tested are described below in each assessment and in the online SI.  The range of performance values observed over all experiments is summarized in Table 6.

\begin{table}[htpb]
    \small
    \centering
    Correct Rates by Question Type and Assessment
    \begin{tabular}{|c|c|c|c|}
        \hline
        \textbf{Assessment} & \textbf{Amount} & \textbf{MCQ} & \textbf{Short Answer} \\ \hline
        \hyperref[sec:realistic_exam]{Assessment 1}  & 5.7 - 9.4\% & 22.3 - 28.1\% & 0\% \\ \hline
        \hyperref[sec:knowledge_application_assessment]{Assessment 2} & N/A & 50.0 - 57.6\% & N/A \\ \hline
    \end{tabular}
    \label{tab:experiment_summary}
    \caption{Correct rates by question type and assessment as measured by all-experiment range of mean prompt performance between Assessment 1 and Assessment 2. Baseline for Multiple Choice is 22.67\% for Assessment 1, 25\% for Assessment 2.  Description of best prompts and parameters is provided below and prompt details are available in SI.}
\end{table}

\subsection*{Assessment 1}
As expected, the quantitative reasoning and arithmetic required in Assessment 1 resulted in substantially lower zero-shot performance than observed in Assessment 2.  Out of 24 questions that required the test-taker to provide a numeric answer based on facts and work papers, GPT-3.5 frequently only answered one, two, or three questions correctly, resulting in an average range across all parameters and prompts of 5.7 to 9.4\%.  While it is arguable whether 0\% is the true baseline for this task, it is clear that such zero-shot performance is not on par with human test-takers.

GPT-3.5 also struggled with arithmetic on the 15 MCQs on Assessment 1, scoring above random chance for some, but not all, prompts and parameters.  As a number of questions include more than four choices, the true baseline rate of guessing is 22.67\%, not 25\%, but despite this, the best prompts and parameters were only 4-6\% above the baseline rate.

Based on a qualitative review of these questions and the model's responses, we believe that performance could be improved somewhat in few-shot evaluations.  Further, we believe that even some zero-shot performance improvements could be achieved by expanding the prompt to include ``scratchpads'' for common relationships or equations \cite{zelikman2022}, as might be seen on problems that feature common workpapers like a statement of cash flows; however, in this paper, we focus on a zero-shot, ``out-of-the-box'' evaluation, and so these improvements are left for future research.

\subsection*{Assessment 2}
As discussed in \hyperref[sec:knowledge_application_assessment]{Assessment 2}, we created 208 MCQs for Assessment 2 to evaluate GPT-3.5's capabilities at the foundation of knowledge work.  Each of these 208 questions has four options, and therefore, the baseline guessing rate for the model is exactly 25\%.  We assessed GPT-3.5 on 208-question assessment exactly 180 times - three samples for each combination of 10 prompts, three \textsc{temperature} ($T$) values, and two \textsc{best\_of} ($n$) parameter values ($3 \cdot 10 \cdot 3 \cdot 2$).  Across these 10 prompts, mean performance ranged between 51.1\% and 56.9\%, with a worst run of 50.0\% (Prompt 13, $T=1.0$) and a best run of 57.6\% (Prompt 16, $T=0.0$).  We did not find significant differences between $n$ parameter values in this assessment.

\begin{table}[htpb]
    \small
    \centering
    \begin{tabular}{|c|c|c|c|}
    \hline
    \textbf{Section} & \textbf{Accuracy} & \textbf{Accuracy - Top Two}\\\hline 
    AUD & 57.1\% & 84.9\%\\\hline
    BEC & 69.7\% & 85.7\%\\\hline
    FAR & 51.0\% & 82.4\%\\\hline
    REG & 53.1\% & 75.8\%\\\hline
    \end{tabular}
    \caption{Accuracy of GPT-3.5 by section of AICPA Exam Blueprints for best prompt and parameter, with correct rate including second-best answer in parentheses. Passage rates are provided in Table \ref{tab:student_performance_by_section} below for reference, but should not be directly compared with model accuracy rates for the reasons discussed above.}
    \label{tab:model_performance_by_section}
\end{table}

\begin{figure}[h!]
    \centering
    \includegraphics[width=5in]{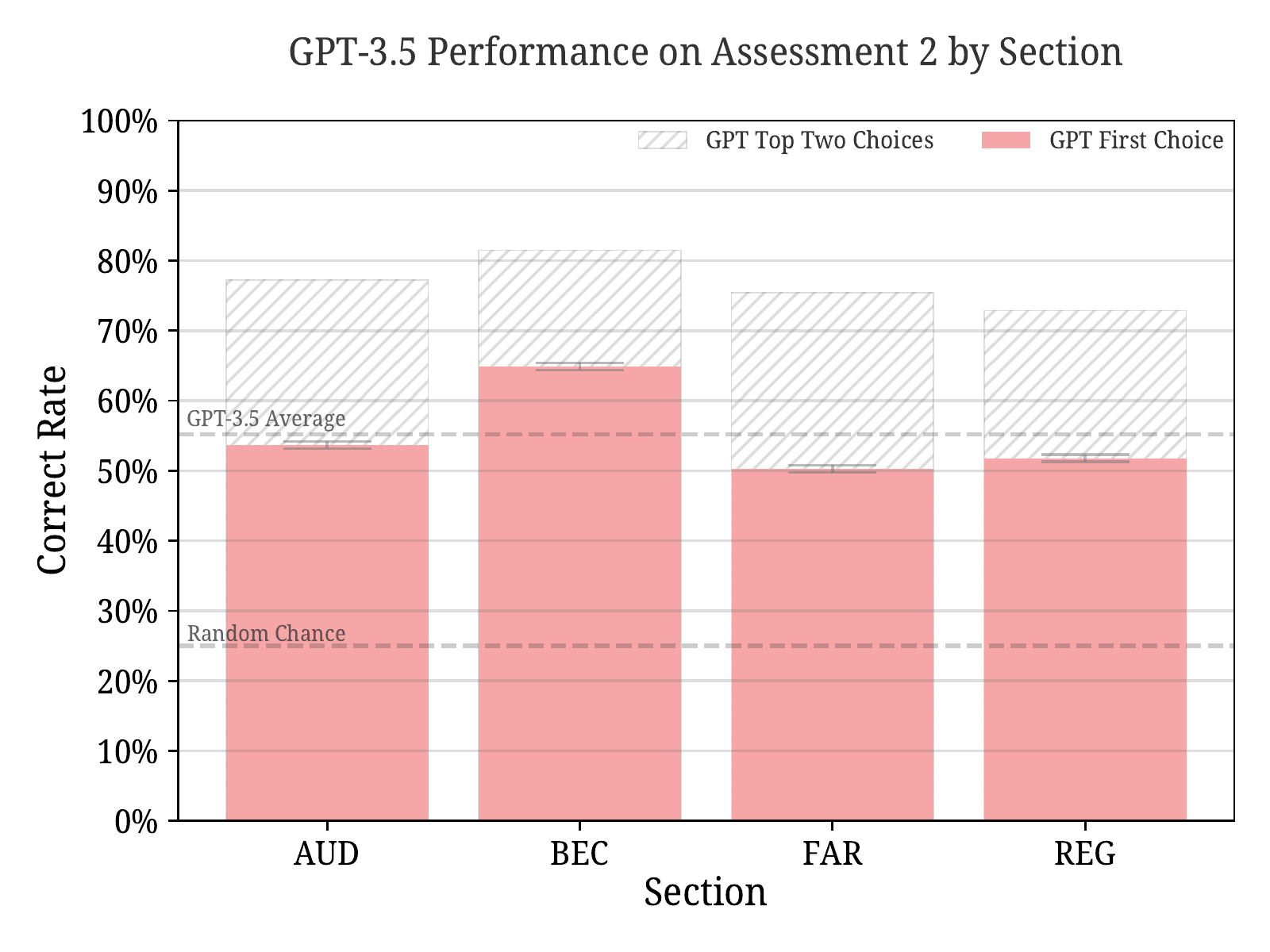}
    \caption{Performance of GPT-3.5 by section of AICPA Exam Blueprints for best prompt and parameter, with correct rate including second-best 
answer in dashed region. Error bars are $\pm1$ standard error of the mean. Note that GPT-3.5 is not assessed on Analysis or Evaluation tasks, unlike human test-takers, and that the percentage of questions correct does not scale linearly with score or passage.}
    \label{fig:best_model_performance}
\end{figure}

Table \ref{tab:model_performance_by_section}, Table \ref{tab:student_performance_by_section}, and Figure \ref{fig:best_model_performance} show the performance of this best prompt and parameter value, including the average percentage of correct questions by section and the average passage rate for test-takers in 2022 as reported by \cite{aicpa2023homepage}.  Overall, GPT-3.5 is demonstrating performance significantly in excess of guessing, achieving approximately 70\% in questions on Business Environment and Concepts (BEC), 57\% for Auditing and Attestation (AUD), 53\% for Regulation (REG), and 51\% for Financial Accounting and Reporting (FAR).  Furthermore, as seen in prior research \cite{bommarito2022gpt}, GPT-3.5 demonstrates strong non-entailment performance as represented by its rank ordering of choices.  The model's top two answers are correct over 82\% of the time, significantly in excess of the 50\% baseline.

While we did not qualitatively code all 208 question for the applicable AICPA skill level, we did review all 53 questions from the Regulation section in Assessment 2.  We found that at least 23 of the 53 questions ($\approx$43\%) require some degree of Application or Analysis.  While these skill levels may be subjective in the context of realistic questions, we encourage readers to examine the complete set of 208 questions in the SI for themselves and to self-assess their own performance to set expectations regarding task type and difficulty.

We do not have a head-to-head comparison between real test-takers and GPT-3.5 for Assessment 2.  Based on our experience, however, we believe that these questions are at least as difficult as the real Remembering \& Understanding and Application questions on the Exam.  Further, the tasks tested in Assessment 2 also account for the vast majority of tasks and types of tasks covered in the AICPA Blueprints.  In addition to reviewing models for single correct answers, some prompts also required models to provide explanations or justifications.  We performed a qualitative review of explanations and justifications for a sample of sessions, and found that more than half of the model's correct answers were also correctly explained with the correct reference or authority.  Interested readers are directed to the online SI for thousands of examples of responses from the model.  Out of all explanations, including incorrect ones, explanations included at least one hallucinated reference or authority in approximately 37\% of the time.  Research is ongoing on the optimal degree of hallucination and techniques for mitigating unwanted hallucination \cite{ji2022hallucinating}, and we will continue to explore these questions and applications in future work.

 \onecolumn

 \begin{figure}[h!]
    \centering
    \includegraphics[width=5in]{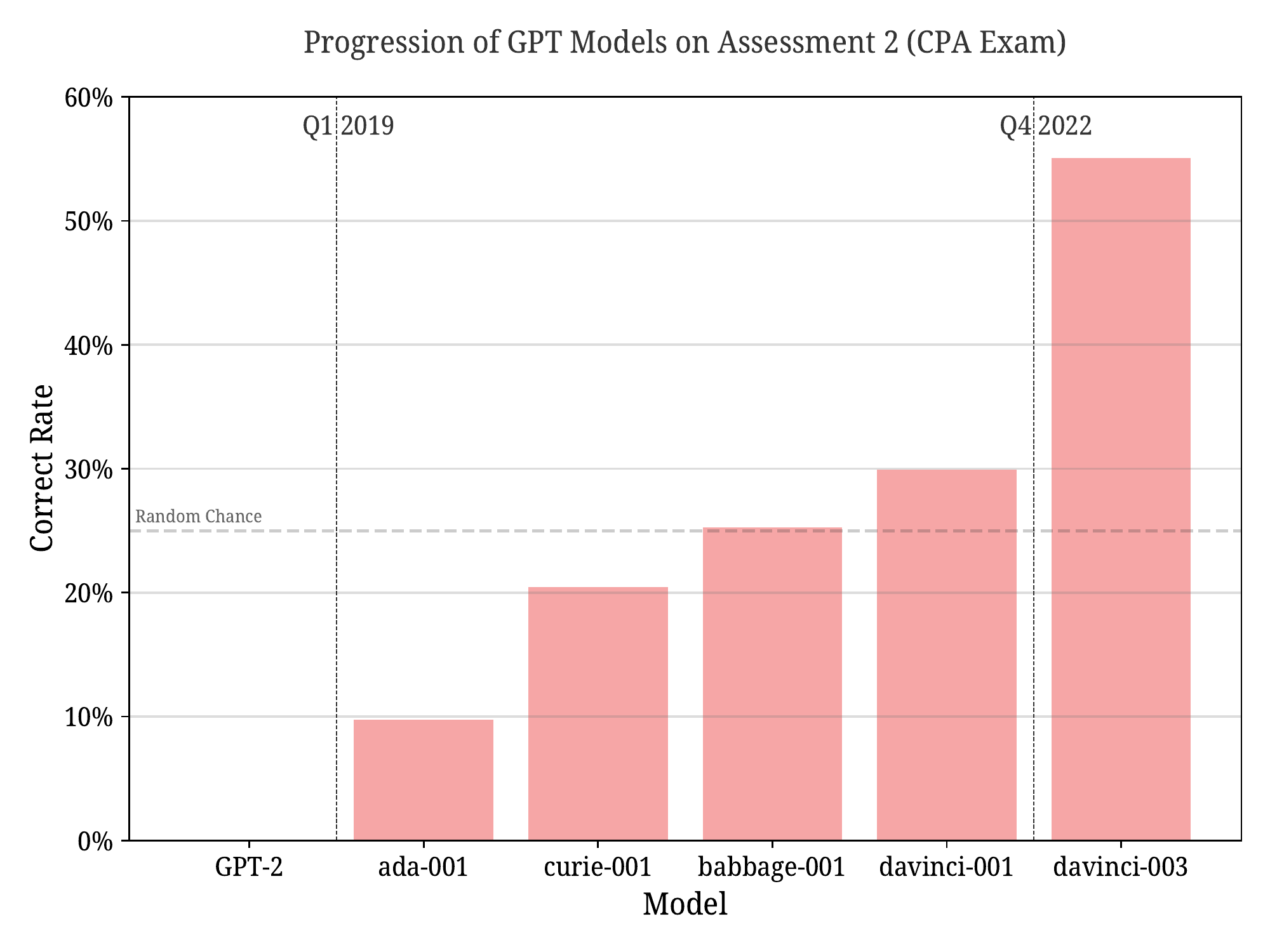}
    \caption{Comparison of model performance across GPT-3 generations.  For \textsc{text-davinci-003}, the average is reported across all runs; for other models, a subset of representative prompts and parameters were included.  GPT-2 was unable to reliably respond to the prompt as instructed and questions were larger than its maximum input token length. More details are available in source and data in the online SI.}
    \label{fig:model_progression}
\end{figure}

\twocolumn

\begin{table}[htbp]
    \small
    \centering
    \begin{tabular}{|c|c|}
        \hline
        \textbf{Model} & \textbf{Correct}\\\hline
        \textsc{text-davinci-003} & 55.1 \\\hline
        \textsc{text-davinci-001} & 29.9 \\\hline
        \textsc{text-babbage-001} & 25.2 \\\hline
        \textsc{text-curie-001} & 20.4 \\\hline
        \textsc{text-ada-001} & 9.7 \\\hline
    \end{tabular}
    \caption{Comparison of model performance across GPT-3 generations.  For \textsc{text-davinci-003}, the average is reported across all runs; for other models, a subset of representative prompts and parameters were included.  More details are available in source and data in the online SI.}
    \label{tab:performance_by_model_generation}
\end{table}

\subsection*{GPT Model Progression}
In prior work \cite{bommarito2022gpt}, we noted that \textsc{text-davinci-003} demonstrated material improvements from prior generations of GPT models.  In this work, we also compare our results against  older or smaller GPT-3 models.  Table \ref{tab:performance_by_model_generation} 
 and Figure \ref{fig:model_progression} summarize these findings, demonstrating a qualitatively-identical story from our work on the Bar Exam.  Only \textsc{text-davinci-001} exhibits the ability to follow instructions and answer above random chance, and between 001 and 003, the spread over random guessing has increased from less than 5\% to over 30\%.
 
\section*{Conclusion and Future Work}
In this paper, we document and develop two assessments of knowledge worker readiness based on the AICPA's Uniform CPA Examination Blueprints.  Assessment 1 is a sample Regulation test as provided by the AICPA, including quantitative reasoning and calculations; Assessment 2 covers foundational skill levels, excluding quantitative reasoning and calculations, for all four sections of the Blueprints.  In total, these assessments cover a broad, practical curriculum including law, finance, accounting, and technology.  We then experimentally evaluate GPT-3.5 on these two assessments, including detailed steps to replicate this evaluation, and share source code and data for all questions not covered by copyright.

First, we find that \textsc{text-davinci-003} achieves a correct rate of 14.4\% on Assessment, significantly underperforming test-takers. As many authors have documented in research on large language models \cite{qian2022limitations, sharma2022overcoming, muffo2022evaluating}, arithmetic and quantitative reasoning are often outside the scope of zero-shot use cases, and these results are consistent with these prior findings.  

As arithmetic and quantitative reasoning are the subjects of substantial active research, we look forward to exploring zero-shot approaches as new models or techniques become available.  Further, as many industrial applications will support iterative or few-shot approaches, we are continuing to investigate applied use cases like the calculation of financial or operational metrics or the analysis of specific financial statements 
using more mature techniques like \cite{nye2021}.

Second, we find that \textsc{text-davinci-003} can achieve an accuracy of 57\% on Assessment 2, significantly better than a 25\% guessing rate, and approaching or on par with anecdotal test-taker performance.  It also demonstrates strong non-entailment capabilities and improving explanation capabilities, as its top two answers are correct 82\% of the time and explanations are correct more often than not.  While this assessment is not identical to the CPA Exam and the AICPA does not publish directly comparable statistics, approximately 45-55\% of test-takers fail the exams annually, as an indication of general difficulty.  All questions in this assessment are available for readers to review and self-assess, and we encourage others to suggest improvements or perform their own assessment on this material.

Finally, as in prior research, we find that recent generations of GPT-3 demonstrate material improvements on this assessment.  While \textsc{text-ada-001} could barely follow instructions and \textsc{text-davinci-001} only exceeded random chance by 5\%, \textsc{text-davinci-003} is now approaching human performance on this assessment.  

As organizations and institutions around the world depend on knowledge workers to navigate an increasingly complex legal and financial landscape \cite{ruhl2017harnessing, bommarito2017measuring}, it is critical that we develop tools that can help safely, effectively meet this demand for knowledge work.  Our findings strongly suggest that future large language models have the potential to transform the quality and efficiency of knowledge work at least as much as search engines did at the turn of the 21st century.  

\section*{Acknowledgments}
Although the original draft of this paper was written by the authors, portions of this paper were fine-tuned by \textsc{text-davinci-003} for formatting and clarity.

\section*{Supplementary Information}
Almost all of the material used in the creation and presentation of this research is available in the online Supplementary Information (SI) at the following URL:\\
\small \url{https://github.com/mjbommar/gpt-as-knowledge-worker}.

 \bibliographystyle{elsarticle-num} 
 \bibliography{references}





\end{document}